%% file: ms.tex
\documentclass[11pt,letterpaper]{article}
\usepackage[letterpaper]{geometry}
\usepackage{emnlp2018}
\usepackage{times}
\usepackage{latexsym}
\usepackage{amsmath}
\usepackage{multirow}
\usepackage{url}
\usepackage{flushend}
\makeatletter
\aclfinalcopy
\newcommand{\@emptybiblabel}[1]{}
\makeatother

\usepackage{graphicx}
\usepackage{CJKutf8}

\title{Sanskrit Sandhi Splitting using $\pmb{seq2(seq)^2}$}
\author{Rahul Aralikatte \\
  IBM Research \\\And
  Neelamadhav Gantayat \\
  IBM Research \\
  {\tt \{rahul.a.r, neelamadhav, naveen.panwar\}@in.ibm.com} \\\And
  Naveen Panwar \\
  IBM Research \\\AND
  Anush Sankaran \\
  IBM Research \\
  {\tt anussank@in.ibm.com} \\\And
  Senthil Mani \\
  IBM Research \\
  {\tt sentmani@in.ibm.com} \\
}

\date{}

\begin{document}
\maketitle
\begin{abstract}
In Sanskrit, small words (morphemes) are combined to form compound words through a process known as \textit{Sandhi}. Sandhi splitting is the process of splitting a given compound word into its constituent morphemes. Although rules governing word splitting exists in the language, it is highly challenging to identify the location of the splits in a compound word. Though existing Sandhi splitting systems incorporate these pre-defined splitting rules, they have a low accuracy as the same compound word might be broken down in multiple ways to provide syntactically correct splits.
	
In this research, we propose a novel deep learning architecture called Double Decoder RNN (DD-RNN), which (i) predicts the location of the split(s) with $95\%$ accuracy, and (ii) predicts the constituent words (learning the Sandhi splitting rules) with $79.5\%$ accuracy, outperforming the state-of-art by $20\%$. Additionally, we show the generalization capability of our deep learning model, by showing competitive results in the problem of Chinese word segmentation, as well.
	
	% In this research, we formulate a Sandhi split location prediction algorithm for compound Sanskrit words using a deep bi-directional character RNN with Attention(AB-RNN). The proposed AB-RNN takes a compound word as input in the form of a character sequence and predicts the probability of split at character index. The advantage of the proposed AB-RNN is its ability to remember long term dependencies in a given sequence, thus providing good performance. We further release a large scale corpus for evaluating sandhi splitting techniques and benchmark our technique on it.
	
	% To the best of our knowledge, deep learning techniques have never been applied to the Sanskrit Sandhi splitting problem before.
\end{abstract}

%\section{Credits}
%The authors would like to thank \emph{Shubham Bharadwaj}, for his contribution towards creating manual corpus.

\input{1_Introduction}
\input{5_proposed}
\input{3_existing_new}

\input{6_evaluation}

\input{7_conclusion}

%\bibliography{acl2017}
\bibliography{main}
\bibliographystyle{acl_natbib_nourl}

%An ancient scholar, \textit{P\={a}\d{n}ini}, is believed to have created the underlying grammar for Sanskrit Sandhi process, which form the basis of Indo-Aryan languages and further used in most of the tonal language currently in practice.

\end{document}

\maketitle
	
\appendix

\section{Sandhi splitting challenges}
Some of the major challenges faced by existing Sandhi splitting tools are briefly described below to motivate the hardness of the problem:

\begin{figure}[]
	\centering
	\includegraphics[width=0.5\textwidth]{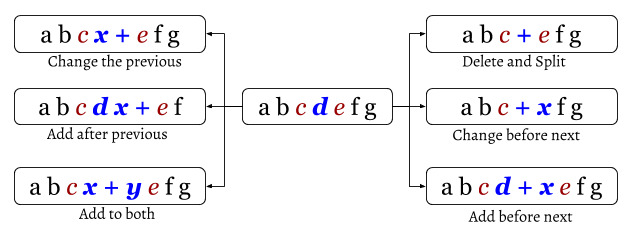}
	\caption{\label{exmotive}An example illustrating the different kinds of syntactical splits and the challenges for a sequence learning algorithm.}
\end{figure}

\begin{enumerate}
	\itemsep0em
	\item \textbf{Identifying multiple locations of split:} Identifying the location in a word where split has to be performed is the most challenging problem in performing splitting. As shown in Figure~\ref{exmotive}, transformation can happen in any location and in any form. Further, sandhi splitting involves identifying multiple potential locations, and validating them based on the previous locations. 
	\item \textbf{Cascading split effect:} There are some rules in which the effect of a split is not merely restricted to the immediate vicinity (neighboring characters). For example, in \textit{uttar\={a}ya\d{n}a} $\rightarrow$ \textit{uttara} + \textit{ayana}, the \textit{r} of \textit{uttara} changes the \textit{\d{n}} of \textit{ayana} to \textit{n}. 
	\item \textbf{Sam\={a}sa:} The process of \textit{Sam\={a}sa} is a process similar to Sandhi where words come together by discarding majority of their characters. A subset of the rules governing \textit{Sam\={a}sa} overlaps with Sandhi. Thus, Sandhi splitters need to maintain two rule sets to correctly identify the constituent words. Existing systems require the user to explicitly pass the intermediate results back to it to perform the splitting correctly. For example, existing systems correctly split the word with a \textit{Sam\={a}sa} \textit{lak\d{s}yasy\={a}rthatvavyavah\={a}r\={a}nurodhena} to form \textit{lak\d{s}yasya} + \textit{arthatvavyavah\={a}ra} + \textit{anurodhena}. However the second word, \textit{arthatvavyavah\={a}ra}, contains a Sandhi and must be sent back into the system to get its constituent words.
	\item \textbf{Incomplete rule set:} Though most of the splitting rules can easily be identified, there are many nuances which are often difficult to handle. There are also some rules which occur very rarely. For example, \textit{sa yog\={\i}} $\rightarrow$ \textit{sa\d{h}} + \textit{yog\={\i}}. Incomplete rule set during splitting will result in false negatives, such as, none of the existing splitters split (\textit{a} + \textit{chedya\d{h}} $\rightarrow$ \textit{acchedya\d{h}}), correctly as \textit{a} may not have the associated rule captured. Thus, heuristically defining all the splitting rules will be intractable, while learning them from examples is more generalizable.
\end{enumerate}

\iffalse
\section{Implementation details}
The architecture of the DD-RNN is shown in Figure~\ref{fig:architecture}. We used a character embedding size of $128$. The bi-directional encoder and the two decoders are $2$ layers deep with $512$ LSTM units in each layer. A dropout layer with $p=0.3$ is applied after each LSTM layer. The entire network is implemented in Torch~\footnote{\url{http://torch.ch/}}.

Of the $71,747$ words in our benchmark dataset, we randomly sampled $80\%$ of the data for training our models. The remaining $20\%$ was used for testing. We used stochastic gradient descent for optimizing the model parameters with an initial learning rate of $1.0$. The learning rate was decayed by a factor of $0.5$ if the validation perplexity did not improve after an epoch. We used a batch size of $64$ and trained the network for $10$ epochs on four Tesla K80 GPUs. This setup remains the same for all the experiments we conduct.

\begin{figure*}[]
	\centering
	\includegraphics[width=\textwidth]{images/architecture.png}
	\caption{\label{fig:architecture}The bi-directional encoder and decoders with attention}
\end{figure*}
\fi

%% file: 1_Introduction.tex
\section{Introduction}
Compound word formation in Sanskrit is governed by a set of deterministic rules following a well-defined structure described in \textit{P\={a}\d{n}ini's} \textit{A\d{s}\d{t}\={a}dhy\={a}y\={\i}}, a seminal work on Sanskrit grammar. The process of merging two or more morphemes to form a word in Sanskrit is called  \textit{Sandhi} and the process of breaking a compound word into its constituent morphemes is called \textit{Sandhi splitting}. In Japanese, \textit{Rendaku} (`sequential voicing') is similar to Sandhi. For example, \textit{`origami'} consists of \textit{`ori'} (paper) + \textit{`kami'} (folding), where \textit{`kami'} changes to \textit{`gami'} due to \textit{Rendaku}.

Learning the process of sandhi splitting for Sanskrit could provide linguistic insights into the formation of words in a wide-variety of Dravidian languages. From an NLP perspective, automated learning of word formations in Sanskrit could provide a framework for learning word organization in other Indian languages, as well~\cite{bharati2006building}. In literature, past works have explored sandhi splitting~\cite{Gillon2009}~\cite{kulkarni2009sanskrit}, as a rule based problem by applying the rules from \textit{A\d{s}\d{t}\={a}dhy\={a}y\={\i}} in a brute force manner.
\begin{figure}[t]
	\center
	\includegraphics[scale=0.35]{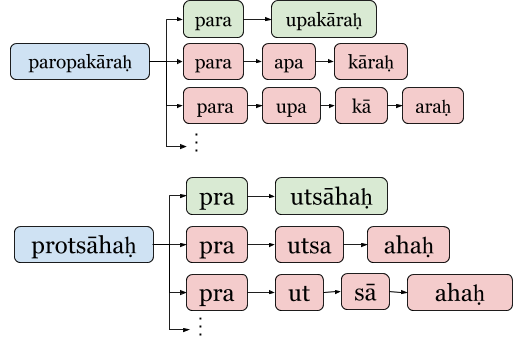}
	\caption{Different possible splits for the word \textit{paropak\={a}ra\d{h}} and \textit{prots\={a}ha\d{h}}, provided by a standard Sandhi splitter.}
	\label{fig:ex2}
\end{figure}
Consider the example in Figure~\ref{fig:ex2} illustrating the different possible splits of a compound word \textit{paropak\={a}ra\d{h}}. While the correct split is \textit{para} + \textit{upak\={a}ra\d{h}}, other forms of splits such as, \textit{para} + \textit{apa} + \textit{k\={a}ra\d{h}} are syntactically possible while semantically incorrect\footnote{Different syntactic splits given by one of the popular Sandhi splitters: \url{https://goo.gl/0M5CPS} and \url{https://goo.gl/JHnpJw}}. Thus, knowing all the rules of splitting is insufficient and it is essential to identify the location(s) of split(s) in a given compound word. 

% On an abstract level, the Chinese word segmentation problem looks very similar to Sandhi splitting. But in Chinese, when words combine, they do not change morphologically. The $n^{th}$ morpheme is just appended to the $(n-1)^{th}$ morpheme. For example: 
% \begin{CJK*}{UTF8}{bsmi}致以 + 亲切 + 的 + 问候\end{CJK*} $\rightarrow$ \begin{CJK*}{UTF8}{bsmi}致以亲切的问候\end{CJK*}. 
% Thus, the Chinese segmentation problem boils down as a special case of sandhi splitting. LSTMs and Gated Recursive Neural Networks(GRNNs) have been applied to this problem by \cite{chinLSTM} and \cite{chingated} respectively. We compare our model with these approaches and show the effectiveness of the proposed model in Section~\ref{eval}.

%\cite{chindepgated} uses a dependency based GRNN which expects a dependency tree as input to predict the split locations. Therefore this technique cannot be applied to Sanskrit as no dependency parser exists for the language. The technique mentioned in \cite{chinLSTM} can only predict split locations and not the split words themselves. Therefore we can only compare our split location prediction with this technique which we do in Section~\ref{eval}.  

In this research, we propose an approach for automated generation of split words by first learning the potential split locations in a compound word. We use a deep bi-directional character RNN encoder and two decoders with attention, $\pmb{seq2(seq)^2}$. %AB-RNN considers the compound word as a sequence of characters and provides a probability of split at each location of the sequence. 
%We curate and release to the public domain a large-scale benchmark dataset and evaluate our approach on that. The dataset contains $71,747$ compound words and their ground truth splits. 
The accuracy of our approach on the benchmark dataset for split location prediction is $95\%$ and for split words prediction is $79.5\%$ respectively. To the best of our knowledge, this is the first research work to explore deep learning techniques for the problem of Sanskrit Sandhi splitting, along with producing state-of-art results. Additionally, we show the performance of our proposed model for Chinese word segmentation to demonstrate the model's generalization capability.

%% file: 5_proposed.tex
\section{$\pmb{seq2(seq)^2}$: Model Description} \label{propo}

In this section, we present our double decoder model to address the Sandhi splitting problem. We first outline the issues with basic deep learning architectures and conceptually highlight the advantages of the double decoder model.

\subsection{Issues with standard architectures}
Consider an example of splitting a sequence \textit{abcdefg} as \textit{abcdx} + \textit{efg}. The primary task is to identify \textit{d} as the split location. Further, for a given location \textit{d} in the character sequence, the algorithm should take into account (i) the context of character sequence \textit{abc}, (ii) the immediate previous character \textit{c}, (iii) the immediate succeeding character \textit{e}, to make an effective split. For such sequence learning problems, RNNs have become the most popular deep learning model~\cite{pascanu2013construct}~\cite{sak2014long}. 
% Specifically in the domain of natural language processing, these algorithms have been highly successful and have produced state-of-art performance in language modeling~\cite{hermans2013training} and machine translation~\cite{cho2014learning}. 

A basic RNN encoder-decoder model~\cite{cho-ende} with LSTM units~\cite{hochreiter1997long}, similar to a machine translation model, was trained initially. The compound word's characters is fed as input to the encoder and is translated to a sequence of characters representing the split words (`+' symbol acts as a separator between the generated split words). However, the model did not yield adequate performance as it encoded only the context of the characters that appeared before the potential split location(s). Though we tried making the encoder bi-directional (referred to as B-RNN), the model's performance only improved marginally. Adding global attention (referred to as B-RNN-A) to the decoder enabled the model to attend to the characters surrounding the potential split location(s) and improved the split prediction performance, making it comparable with some of the best performing tools in the literature.

\subsection{Double Decoder RNN (DD-RNN) model}
The critical part of learning to split compound words is to correctly identify the location(s) of the split(s). Therefore, we added a two decoders to our bi-directional  encoder-decoder model: (i) \textit{location decoder} which learns to predict the split locations and (ii) \textit{character decoder} which generates the split words. A compound word is fed into the encoder character by character. Each character's embedding $x_i$ is passed to the encoders LSTM units. There are two LSTM layers which encode the word, one in forward direction and the other backward. The encoded context vector $e_i$ is then passed to a global attention layer. 

In the first phase of training, only the \textit{location decoder} is trained and the \textit{character decoder} is frozen. The character embeddings are learned from scratch in this phase along with the attention weights and other parameters. Here, the model learns to identify the split locations. For example, if the inputs are the embeddings for the compound word \textit{prots\={a}ha\d{h}}, the location decoder will generate a binary vector $[0, 0, 1, 0, 0, 0, 0, 0, 0]$ which indicates that the split occurs between the third and fourth characters.
In the second phase, the \textit{location decoder} is frozen and the \textit{character decoder} is trained. The encoder and attention weights are allowed to be fine-tuned. This decoder learns the underlying rules of Sandhi splitting. Since the attention layer is already pre-trained to identify potential split locations in the previous phase, the \textit{character decoder} can use this context and learn to split the words more accurately. For example, for the same input word \textit{prots\={a}ha\d{h}}, the \textit{character decoder} will generate [p, r, a, +, u, t, s, \={a}, h, a, \d{h}] as the output. Here the character \textit{o} is split into two characters \textit{a} and \textit{u}.

\begin{figure*}[]
	\centering
	\includegraphics[width=\textwidth]{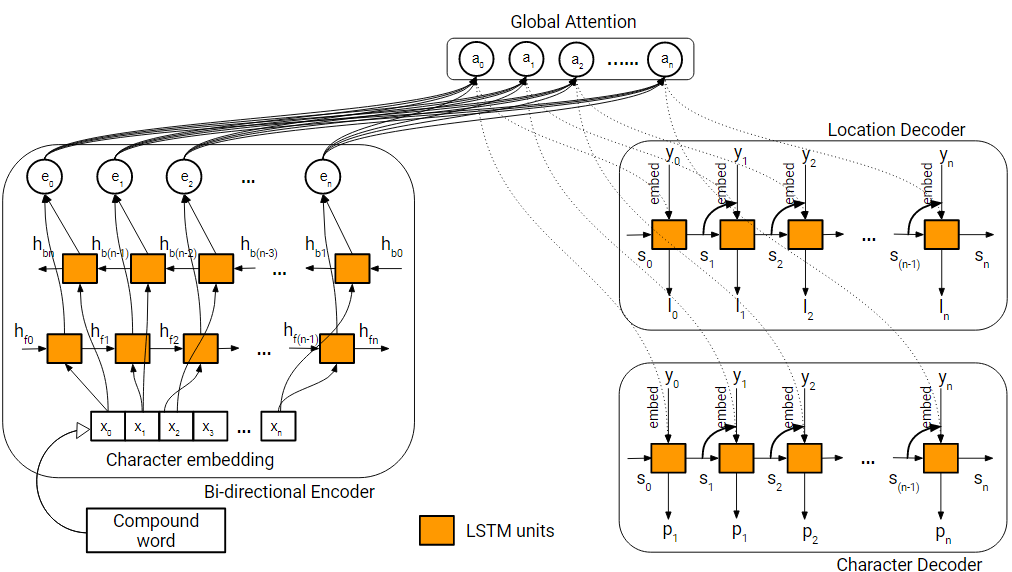}
	\caption{\label{fig:architecture}The bi-directional encoder and decoders with attention}
\end{figure*}

In both the training phases, we use negative log likelihood as the loss function. Let $X$ be the sequence of the input compound word's characters and $Y$ be the binary vector which indicates the location of the split(s) in the first phase and the true target sequence of characters which form the split words in the second phase. If $Y = {y_1, y_2, ..., y_n}$, then the loss function is defined as:
\[loss = -\sum_{i=1}^{|Y|} \log{P(y_i | y_{i-1}, \cdots , y_1, X)}\]

We evaluate the DD-RNN and compare it with other tools and architectures in Section~\ref{eval}.

\subsection{Implementation details}
The architecture of the DD-RNN is shown in Figure~\ref{fig:architecture}. We used a character embedding size of $128$. The bi-directional encoder and the two decoders are $2$ layers deep with $512$ LSTM units in each layer. A dropout layer with $p=0.3$ is applied after each LSTM layer. The entire network is implemented in Torch~\footnote{\url{http://torch.ch/}}.

Of the $71,747$ words in our benchmark dataset, we randomly sampled $80\%$ of the data for training our models. The remaining $20\%$ was used for testing. We used stochastic gradient descent for optimizing the model parameters with an initial learning rate of $1.0$. The learning rate was decayed by a factor of $0.5$ if the validation perplexity did not improve after an epoch. We used a batch size of $64$ and trained the network for $10$ epochs on four Tesla K80 GPUs. This setup remains the same for all the experiments we conduct.

\iffalse
We summarize the reasoning behind the various components in our model as follows:
\begin{itemize}
		\itemsep0em
	\item We use two decoders which essentially split the learning process between them. The \textit{location decoder} allows the model to learn the locations of the splits whereas the \textit{character decoder} generates the split words.
	
	\item The encoder is bidirectional~\cite{graves2013speech} which considers the input sequence in both forward (first to last) and backward (last to first) directions and merges them to generate the encoding. Thus, the context of a character \textit{d} includes both the preceding (\textit{abc}) and succeeding characters (\textit{efg}). 
	
	\item The attention mechanism~\cite{luong2015effective} is used with the decoders. In the first phase of training, it learns to attend to certain portions of the word where the splits are most likely to occur. In the second phase, it helps the \textit{character decoder} to apply to splitting rules at the correct indexes. % Also, compound words in Sanskrit can be long and can have more than $50-60$ characters. The longest word in Sanskrit has $1421$ characters. Attention helps us to learn and remember which portions of the sequence are important.
\end{itemize}
\fi

%% file: 3_existing_new.tex
\section{Existing Datasets and Tools} \label{relatmotiv}
In this section, we briefly introduce various Sanskirt Sandhi datasets and splitting tools available in literature. We also discuss the tools' drawbacks and the major challenges faced while creating such tools.

\textbf{Datasets:} 	The \textit{UoH corpus}, created at the University of Hyderabad\footnote{Available at: \url{http://sanskrit.uohyd.ac.in/Corpus/}} contains $113,913$ words and their splits. This dataset is noisy with typing errors and incorrect splits. The recent \textit{SandhiKosh corpus}~\cite{sandhikosh} is a set of $13,930$ annotated splits. We combine these datasets and heuristically prune them to finally get $71,747$ words and their splits. The pruning is done by considering a data point to be valid only if the compound word and it's splits are present in a standard Sanskrit dictionary~\cite{monier1970sanskrit}. We use this as our benchmark dataset and run all our experiments on it.

\textbf{Tools:} There exist multiple Sandhi splitters in the open domain such as (i) \textit{JNU splitter}~\cite{sachin2007sandhi}, (ii) \textit{UoH splitter}~\cite{kk} and (iii) \textit{INRIA sanskrit reader companion}~\cite{huet2003}~\cite{goyal2013completeness}. Though each tool addresses the splitting problem in a specialized way, the general principle remains constant. For a given compound word, the set of all rules are applied to every character in the word and a large potential candidate list of word splits is obtained. Then, a morpheme dictionary of Sanskrit words is used with other heuristics to remove infeasible word split combinations. However, none of the approaches address the fundamental problem of identifying the location of the split before applying the rules, which will significantly reduce the number of rules that can be applied, hence resulting in more accurate splits.

\iffalse
\textbf{Challenges:} Some of the challenges faced when building a Sandhi splitter are as follows:
\begin{enumerate}
	\itemsep0em
	\item \textit{Identifying multiple locations of split:} As mentioned previously, finding the location of a split in a word is one of the most challenging problems of Sandhi splitting. This is made more complex if there are multiple splits in a single word because the $i^{th}$ split depends on the validity of the $i-1^{th}$ split and so on
	\item \textit{Cascading split effect:} There are some rules of Sandhi in which the effect of a split is propagated beyond the immediate vicinity of the split location
	\item \textit{Distinction from Sam\={a}sa:} The \textit{Sam\={a}sa} is a process similar to Sandhi where words come together by discarding majority of their characters. Although they are two independent processes, if not defined accurately, the splitters often apply the rules of Sandhi at places where \textit{Sam\={a}sa} should have been applied
	\item \textit{Confusing rule set:} Though most of the splitting rules can easily be identified, there are many nuances which are often difficult to handle. There are also some rules which occur very rarely in place of other standard rules
\end{enumerate}
\fi

%% file: 6_evaluation.tex
\section{Evaluation and Results} \label{eval}
%In this section, we evaluate the performance of sandhi splitting task using our DD-RNN model with existing splitting tools. Please refer Figure~\ref{fig:graph0} for a comparison of these tools with DD-RNN on our benchmark dataset. From the graph it is evident that the precision of DDRNN is 38.8\% better than the best tool i.e. INRIA.  %It is not possible to directly compare our proposed approach with these splitting tools as we do not perform the actual splitting, but only predict the location of all the splits. But, since sandhi location prediction is a fundamental step in the sandhi splitting process, having a system with very high precision in identifying split locations will naturally lead to a system which can perform sandhi splitting with greater accuracy.
%Figure~\ref{fig:graph2} refers to the precession on the prediction of exact split given the length of the word. From the graph it is evident that with increase of word length precision decreases.  Other tools which splits sandhi are performing worse, as length increases complexity of identifying split location increases, which forces the tools to do wrong splittings.

%\begin{figure}[t]
%	\centering
%	\includegraphics[scale=0.45]{images/accuracy.png}
%	\caption{Precision of different techniques on test dataset including DDRNN}
%	\label{fig:graph0}
%\end{figure}

We evaluate the performance of our DD-RNN model by:
(i) comparing the \textbf{split prediction} accuracy with other publicly available sandhi splitting tools,
(ii) comparing the \textbf{split prediction} accuracy with other standard RNN architectures such as RNN, B-RNN, and B-RNN-A, and
(iii) comparing the \textbf{location prediction} accuracy with the RNNs used for Chinese word segmentation (as they only predict the split locations and do not learn the rules of splitting)

% Further, we show how these metrics vary with input compound word lengths. We also provide some interesting insights about the models.

\subsection{Comparison with publicly available tools}
\begin{figure}[t]
	\centering
	\includegraphics[scale=0.5]{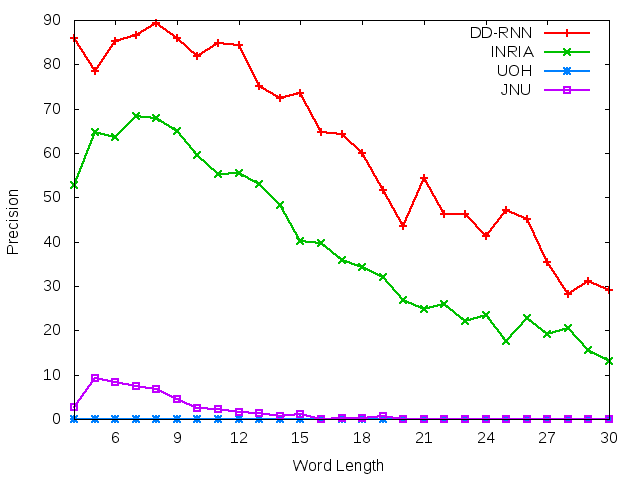}
	\caption{Top-1 split prediction accuracy comparison of different publicly available tools with DD-RNN }
	\label{fig:graph2}
\end{figure}

\begin{figure}[t]
	\centering
	\includegraphics[scale=0.5]{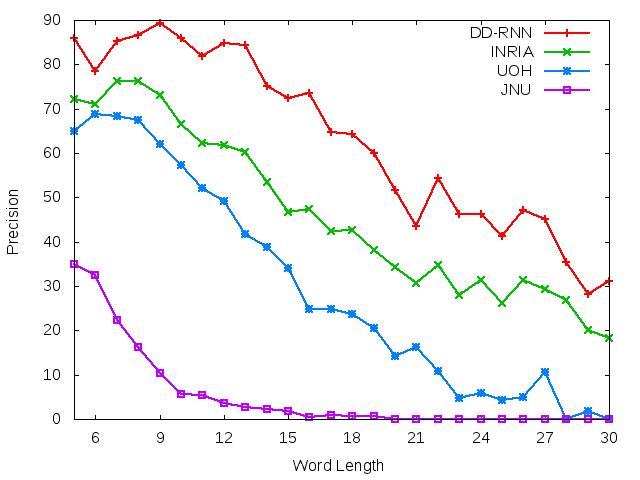}
	\caption{Split prediction accuracy comparison of different publicly available tools (Top-10) with DD-RNN (Top-1)}
	\label{fig:graph3}
\end{figure}

The tools discussed in Section~\ref{relatmotiv} take a compound word as input and provide a list of all possible splits as output (UoH and INRIA splitters provide weighted lists). Initially, we compared only the top prediction in each list with the true output. This gave a very low precision for the tools as shown in Figure~\ref{fig:graph2}. Therefore, we relaxed this constraint and considered an output to be correct if the true split is present in the top ten predictions of the list. This increased the precision of the tools as shown in Figure~\ref{fig:graph3} and Table~\ref{tab7}.

Even though DD-RNN generates only one output for every input, it clearly out-performs the other publicly available tools by a fair margin.

\subsection{Comparison with standard RNN architectures}
To compare the performance of DD-RNN with other standard RNN architectures, we trained the following three models to generate the split predictions on our benchmark dataset:
(i) uni-directional encoder and decoder without attention (RNN),
(ii) bi-directional encoder and decoder without attention (B-RNN), and
(iii) bi-directional encoder and decoder with attention (B-RNN-A)

%\begin{figure}[t]
%	\center
%	\includegraphics[scale=.45]{images/figure_1.png}
%	\caption{Heatmap showing the predicted probability distribution for split location prediction using AB-RNN. Three examples show the prediction of one, two, and three sandhi splits respectively.}
%	\label{fig:diagram2}
%\end{figure}

\begin{table}[]
\centering
\begin{tabular}{r|c|l|c|l|}
\cline{2-5}
\multicolumn{1}{l|}{} & \multicolumn{4}{c|}{\textbf{Accuracy (\%)}} \\ \hline
\multicolumn{1}{|c|}{\textbf{Model}} & \multicolumn{2}{c|}{\textbf{\begin{tabular}[c]{@{}c@{}}Location \\ Prediction\end{tabular}}} & \multicolumn{2}{c|}{\textbf{\begin{tabular}[c]{@{}c@{}}Split \\ Prediction\end{tabular}}} \\ \hline \hline
\multicolumn{1}{|r|}{\textbf{JNU} (Top 10)} & \multicolumn{2}{c|}{-} & \multicolumn{2}{c|}{8.1} \\ \hline
\multicolumn{1}{|r|}{\textbf{UoH} (Top 10)} & \multicolumn{2}{c|}{-} & \multicolumn{2}{c|}{47.2} \\ \hline
\multicolumn{1}{|r|}{\textbf{INRIA} (Top 10)} & \multicolumn{2}{c|}{-} & \multicolumn{2}{c|}{59.9} \\ \hline \hline
\multicolumn{1}{|r|}{\textbf{RNN}} & \multicolumn{2}{c|}{79.10} & \multicolumn{2}{c|}{56.6} \\ \hline
\multicolumn{1}{|r|}{\textbf{B-RNN}} & \multicolumn{2}{c|}{84.62} & \multicolumn{2}{c|}{58.6} \\ \hline
\multicolumn{1}{|r|}{\textbf{B-RNN-A}} & \multicolumn{2}{c|}{88.53} & \multicolumn{2}{c|}{69.3} \\ \hline \hline
\multicolumn{1}{|r|}{\textbf{DD-RNN}} & \multicolumn{2}{c|}{\textbf{95.0}} & \multicolumn{2}{c|}{\textbf{79.5}} \\ \hline \hline
\multicolumn{1}{|r|}{\textbf{LSTM-4}} & \multicolumn{2}{c|}{70.2} & \multicolumn{2}{c|}{-} \\ \hline
\multicolumn{1}{|r|}{\textbf{GRNN-5}} & \multicolumn{2}{c|}{67.7} & \multicolumn{2}{c|}{-} \\ \hline
\end{tabular}
\caption{Location and split prediction accuracy of all the tools and models under comparison}
\label{tab7}
\end{table}

\begin{figure}[t]
	\centering
	\includegraphics[scale=0.45]{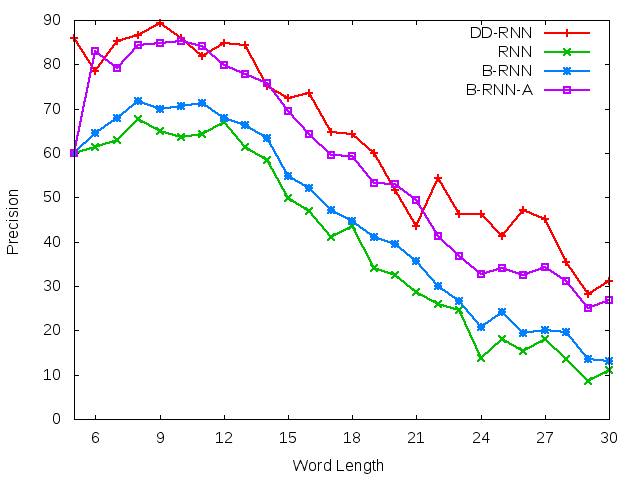}
	\caption{Split prediction accuracy comparison of different variations of RNN on words of different lengths}
	\label{fig:graph1}
\end{figure}

As seen from the middle part of Table~\ref{tab7}, the DD-RNN performs much better than the other architectures with an accuracy of \textbf{79.5\%}. It is to be noted that B-RNN-A is the same as DD-RNN without the \textit{location decoder}. However, the accuracy of DD-RNN is $14.7$\% more than that the B-RNN-A and consistently outperforms B-RNN-A on almost all word lengths (Figure~\ref{fig:graph1}). This indicates that the attention mechanism of DD-RNN has learned to better identify the split location(s) due to its pre-training with the \textit{location decoder}.

\subsection{Comparison with similar works}
\cite{reddy2018} propose a \textit{seq2seq} model with attention to tackle the Sandhi problem. Their model is similar to B-RNN-A and is outperformed by our proposed DD-RNN by \~6.47\%. We also compared our proposed DD-RNN with a uni-directional LSTM with a depth of $4$~\cite{chinLSTM} (\textbf{LSTM-4}) and a Gated Recursive Neural Network with a depth of $5$~\cite{chingated} (\textbf{GRNN-5}). These models were used to get state of the art results for Chinese word segmentation and their source code is made available online.\footnote{\url{https://github.com/FudanNLP}} Since these models can only predict the location(s) of the split(s) and cannot generate the split words themselves, we used the location prediction accuracy as the metric. We trained these models on our benchmark dataset and the results are shown in Table~\ref{tab7}. DD-RNN's precision is $35.3$\% and $40.3$\% better than LSTM-4 and GRNN-5 respectively. Conversely, we trained the DD-RNN for the Chinese word segmentation task to test the generalizability of the model. Since there are no morphological changes during segmentation in Chinese, the character decoder is redundant and the model collapses to simple \textit{seq2seq}. We used the PKU dataset which is also used in~\cite{chinLSTM} \&~\cite{chingated} and obtained an accuracy of 64.25\% which is comparable to the results of other standard models.

\iffalse
We try to provide some intuition as why that is the case in the following paragraphs.

The LSTM-4 model \cite{chinLSTM} is similar to our first basic RNN model. It is four layers deep, but does not follow our encoder-decoder paradigm. Since the RNN is not bi-directional and the whole word is not encoded before making the split location predictions, this model suffers from incomplete information. That is, if the model has to predict whether a split occurs at location $i$, it can do so only by looking at the previous characters (indices $0$ to $i-1$). It does not have information about latter characters (indices $i+1$ to $n$) in the word which may influence the probability of a split. 

The hierarchical GRNN model used in~\cite{chingated} has the ability to capture context from the entire word by recursively building up complex features from the bottom layer to the top. We notice that, this model works very well for smaller words and fails when the word length is long or there are multiple splits. This might again be attributed to the split location identification problem mentioned in the previous sections.
\fi

To summarize, we have used our benchmark dataset to compare the DD-RNN model with existing publicly available Sandhi splitting tools, other RNN architectures and models used for Chinese word segmentation task. Among the existing tools, the INRIA splitter gives the highest split prediction accuracy of 59.9\%. Among the standard RNN architectures, B-RNN-A performs the best with a split prediction accuracy of 69.3\%. LSTM-4 performs the best among the Chinese word segmentation models with a location prediction accuracy of 70.2\%. DD-RNN outperforms all the models both in location and split predictions with 95\% and 79.5\% accuracies, respectively.

%% file: 7_conclusion.tex
\section{Research Impact}
This work can be foundational to other Sanskrit based NLP tasks. Let us consider translation as an example. In Sanskrit, arbitrary number of words can be joined together to form a compound word. Literary works, especially from the \textit{Vedic} era often contain words which are a concatenation of three or more simpler words. Presence of such compound words will increase the vocabulary size exponentially and hinder the translation process. However, as a pre-processing step, if all the compound words are split before training a translation model, the number of unique words in the vocabulary reduces which will ease the learning process.

\section{Conclusion} \label{concl}
In this research, we propose a novel double decoder RNN architecture with attention for Sanskrit Sandhi splitting. 
%Learning such a model would provide further insights into the fundamental linguistic word formation rules of the language. 
A deep bi-directional encoder is used to encode the character sequence of a Sanskrit word. Using this encoded context vector, a \textit{location decoder} is first used to learn the location(s) of the split(s). Then the \textit{character decoder} is used to generate the split words. We evaluate the performance of the proposed approach on the benchmark dataset in comparison with other publicly available tools, standard RNN architectures and with prior work which tackle similar problems in other languages. As future work, we intend to tackle the harder \textit{Samasa} problem which requires semantic information of a word in addition to the characters' context. 
% Further we intend to increase the generalizability of the DD-RNN model to work on word segmentation tasks of other languages. 

%%As future work, we propose the following:
%%\begin{itemize}
%%	\item In addition to just predicting the location of splits, we would like to predict the constituent words themselves and create a complete end to end system for Sanskrit sandhi splitting
%%	\item Modify the AB-RNN so that it can work well on word segmentation tasks of other languages without any modifications
%%\end{itemize}